\documentclass{article}
\usepackage{spconf,amsmath,graphicx}
\usepackage{multicol}
\usepackage{multirow}
\usepackage{enumitem}
\usepackage{amssymb}
\setlist{nosep, leftmargin=14pt}
\usepackage{mwe} 

\title{EVIL: Evidential Inference Learning for Trustworthy Semi-supervised Medical Image Segmentation}
\name{%
\begin{tabular}{@{}c@{}}
Yingyu Chen$^{1}$ \quad 
Ziyuan Yang$^{1}$ \quad 
Chenyu Shen$^{1}$ \quad 
Zhiwen Wang$^{1}$ \quad 
Yang Qin$^{1}$ \quad \\
Yi Zhang$^{2,1}$, Senior Member, IEEE \thanks{Yi Zhang is the corresponding author}
\end{tabular}}

\address{
$^{1}$College of Computer Science, Sichuan University, Chengdu, China\\
$^{2}$ School of Cyber Science and Engineering, Sichuan University, Chengdu, China
}

\begin{document}
\maketitle
\begin{abstract}

Recently, uncertainty-aware methods have attracted increasing attention in semi-supervised medical image segmentation. However, current methods usually suffer from the drawback that it is difficult to balance the computational cost, estimation accuracy, and theoretical support in a unified framework. To alleviate this problem, we introduce the Dempster–Shafer Theory of Evidence (DST) into semi-supervised medical image segmentation, dubbed \textit{\textbf{EV}idential} \textit{\textbf{I}nference} \textit{\textbf{L}earning} (\textit{\textbf{EVIL}}). EVIL provides a theoretically guaranteed solution to infer accurate uncertainty quantification in a single forward pass. Trustworthy pseudo labels on unlabeled data are generated after uncertainty estimation. The recently proposed consistency regularization-based training paradigm is adopted in our framework, which enforces the consistency on the perturbed predictions to enhance the generalization with few labeled data. Experimental results show that EVIL achieves competitive performance in comparison with several state-of-the-art methods on the public dataset.

\end{abstract}
\section{Introduction}
\label{sec:intro}

Medical image segmentation plays an essential role in subsequent clinical or computer-aided diagnosis and fully-supervised learning has achieved great success in the field of automatic image segmentation \cite{ronneberger2015u}. However, annotating medical images is laborious and requires rich professional knowledge \cite{zou2022tbrats}. 

Semi-supervised learning (SSL) has shown great potential to alleviate this problem by leveraging a large set of unlabeled data accompanied with a limited number of labeled data. These methods can be roughly categorized into two types: (1) pseudo-label retraining, which incorporates pseudo labels on unlabeled data for retraining \cite{zoph2020rethinking,feng2020semi,ibrahim2020semi}; and (2) consistency regularization, which enforces the prediction consistency to enhance generalization with various perturbations, such as input perturbation, feature perturbation, and network perturbation \cite{tarvainen2017mean, chen2021semi, ouali2020semi}.

However, since these methods rely heavily on the prediction of pseudo label, false predictions will severely degrade the segmentation performance. To improve the quality of pseudo labels, some uncertainty-aware methods have been proposed, including Monte Carlo dropout (MC-dropout)-based \cite{yu2019uncertainty}, Information-Entropy-based \cite{wanguncertainty}, and Prediction Variance-based \cite{zheng2021rectifying} methods. However, these methods suffer from some problems: (1) Although MC-dropout is mathematically guaranteed by Bayesian theory, its training process is costly due to the multiple sampling operations; (2) Due to the limited sampling times, MC-dropout can't obtain accurate uncertainty quantification; (3) Other two uncertainty estimation methods have advantages in computational cost, but they lack theoretical support, leading to unstable pseudo label generation.

To handle the above issues, we introduce the Dempster–Shafer Theory of Evidence (DST) into semi-supervised medical image segmentation, providing a theoretically guaranteed single-pass solution for uncertainty quantification inference, dubbed \textit{\textbf{EV}idential} \textit{\textbf{I}nference} \textit{\textbf{L}earning} (\textit{\textbf{EVIL}}). Following the training paradigm proposed in \cite{chen2021semi}, EVIL belongs to the consistency regularization method with network perturbation, which imposes the prediction consistency on two networks perturbed with different initialization. In particular, the two networks play different roles. One is a vanilla segmentation network (S-Net) which directly generates the segmentation result. The other network called evidential-network (E-Net) is built from the perspective of DST, which is theoretically guaranteed for reliable predictions. Different from S-Net, the output of E-Net is regarded as the evidence and parameterized into a Dirichlet distribution on segmentation probabilities. Subjective Logic theory (SL) \cite{audun2018subjective} is employed to quantify the predictions and uncertainties of different categories with the Dirichlet distribution in a single inference, which significantly reduces the training time. Then, the trustworthy pseudo labels on unlabeled data are generated. In summary, there are three merits for our proposed EVIL: lower computation cost due to the single-pass operation, accurate uncertainty estimation based on SL and theoretical guarantee based on DST.

The main contributions of this work are summarized as: 1) we introduce DST into SSL and provide a fast accurate uncertainty estimation with theoretical guarantee in a unified framework; 2) a novel network perturbation strategy is proposed, which allows different initialized network optimized with different objectives; and 3) extensive experiments are conducted to validate the effectiveness of our proposed EVIL.

\begin{figure}
    \centering
    \includegraphics[width=\columnwidth]{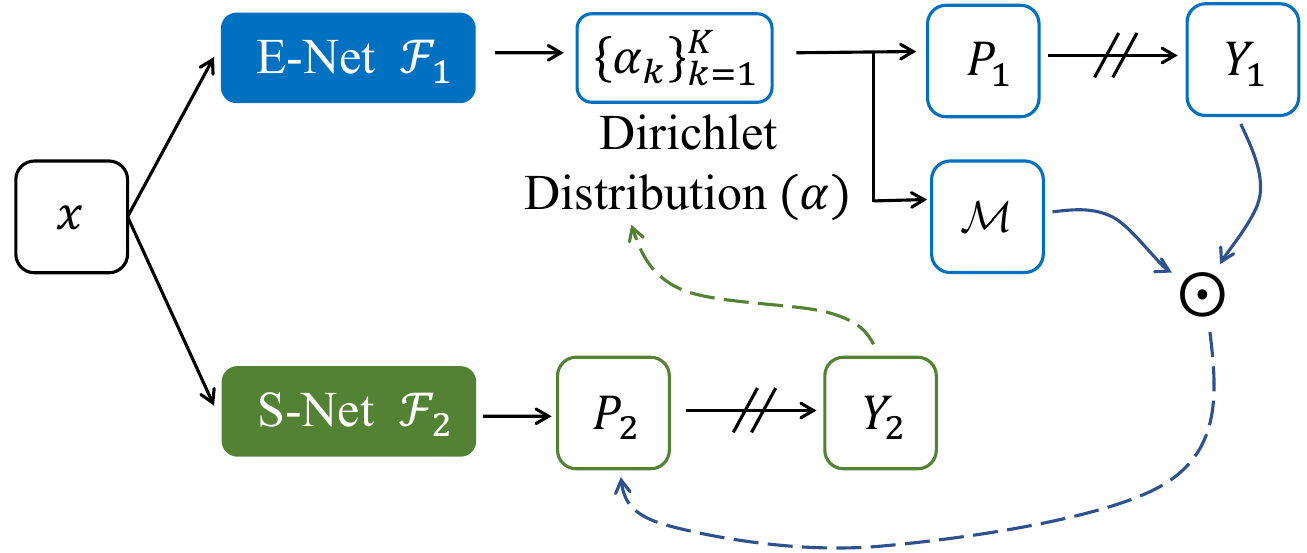}
    \vspace{-0.2cm}
    \caption{The overview of our EVidential Inference Learning framework (EVIL), where $\mathcal{M}$ denotes uncertainty map estimated by E-Net and $\odot$ denotes element-wise product. `$\rightarrow$' presents forward operation, `$\dashrightarrow$' presents supervision loss operation and `$//$' on `$\rightarrow$' presents stop-gradient.}
    \label{fig:my_label}
    \vspace{-0.2cm}
\end{figure}

\section{Method}
Given a labeled set $D_l = \{(\boldsymbol x_i, \boldsymbol y_i)\}^{N_l}_{i=1}$ with ${N_l}$ samples and an unlabeled set $D_u = \{\boldsymbol x_i\}^{N_u}_{i=1}$ with ${N_u}$ samples, where $N_u \gg N_l$ in semi-supervised task. 

As illustrated in Fig. 1, EVIL has two differently initialized networks, E-Net $\mathcal{F}_1$ with parameter set $\theta_1$ and S-Net $\mathcal{F}_2$ with parameter set $\theta_2$.
For labeled data, S-Net is optimized with traditional joint cross-entropy loss and dice loss, while E-Net models a Dirichlet distribution and is optimized with evidential segmentation loss. $P_1$, $P_2$ are the segmentation predictions and $Y_1$, $Y_2$ are the corresponding pseudo labels generated by $argmax$ function. For unlabeled data, E-Net generates pseudo labels $Y_1$ and accurate uncertainty estimations $\mathcal{M}$ simultaneously. 
Then, the trustworthy pseudo labels are calculated by $\mathcal{M} \odot Y_1$ and used to guide the training of S-Net. 
Reversely, the pseudo labels $Y_2$  generated by S-Net is leveraged for E-Net to explore more potential evidence to improve the generation of pseudo labels from unlabeled data.

\subsection{Uncertainty Modeling}

In this section, we utilize DST to model the segmentation uncertainty and generate trustworthy prediction. For a $K$-class segmentation task, given an input $\boldsymbol x_i$, the evidence vector $\boldsymbol e_i$ is obtained with a transform function $g$, which is defined in \cite{Qin2022DECL}:
\begin{equation}
    \boldsymbol e_i = g(\mathcal{F}_1(\boldsymbol x_i)) = exp^{\left(\tanh{\mathcal{F}_1(\boldsymbol x_i)/\tau}\right)},
\end{equation}
where $0 < \tau < 1$ is a scaling parameter set to $1/K$.  $\mathcal{F}_1(\boldsymbol x_i)$ is the output of E-Net with input $\boldsymbol x_i$.
Subjective Logic \cite{audun2018subjective} computes the belief mass for category $k$ and uncertainty as:
\begin{equation}
    b_{i}^k=\frac{e_{i}^k}{ S_i}=\frac{\alpha_{i}^k-1}{ S_i} \quad \text{and} \quad 
    u_i=\frac{K}{S_i},
\end{equation}
where $S_i=\sum_{k=1}^{K}(e_i^k + 1)$, $u_i + \sum_{k=1}^{K} b_i^k=1$ and $\alpha_i^k = e_i^k + 1$. $\boldsymbol{\alpha}_i=\left[\alpha_i^1, \dots, \alpha_i^K \right]$ can be regarded as the parameters of Dirichlet distribution, which models the density of segmentation probability and uncertainty \cite{sensoy2018evidential}. The density function is defined as:
\begin{equation}
    D(\boldsymbol{p}_{i} \mid \boldsymbol{\alpha}_{i}) = 
    \{
    \begin{array}{ll}
        \frac{1}{B(\boldsymbol{\alpha}_i)} \prod_{k=1}^K p_{i}^{\alpha_i^k-1} & \text{for } \boldsymbol{p}_{i} \in \mathcal{S}_{K}, \\
        0 & \text{otherwise},
    \end{array}
\end{equation}
where $\boldsymbol{p}_{i}$ is the segmentation probability, $B(\boldsymbol{\alpha}_i)$ is the $K$-dimensional multinomial beta function for parameter $\boldsymbol{\alpha}_{i}$, and $\mathcal{S}_{K}$ is the $K$-dimensional simplex.

\begin{figure}
\centering
    \centerline{\includegraphics[width=80mm]{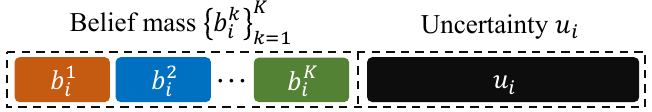}}
    \vspace{-0.2cm}
    \caption{Subjective Logic model, where $u_i + \sum_{k=1}^K b_i^k = 1$.}
    \vspace{-0.4cm}
\end{figure}

\begin{figure*}
    \vspace{-0.2cm}
    \centering
    \begin{minipage}[b]{.138\linewidth} 
      \centering
      \centerline{\includegraphics[width=\linewidth]{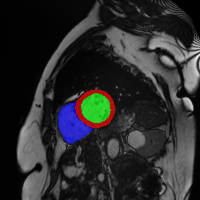}}
    \end{minipage}
    \begin{minipage}[b]{.138\linewidth}
      \centering
      \centerline{\includegraphics[width=\linewidth]{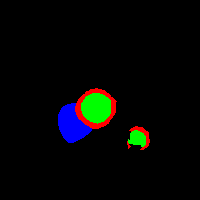}}
    \end{minipage}
    \begin{minipage}[b]{.138\linewidth}
      \centering
      \centerline{\includegraphics[width=\linewidth]{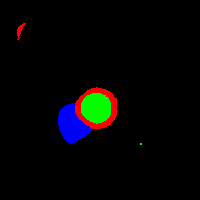}}
    \end{minipage}
    \begin{minipage}[b]{.138\linewidth}
      \centering
      \centerline{\includegraphics[width=\linewidth]{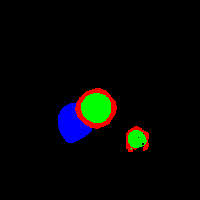}}
    \end{minipage}
    \begin{minipage}[b]{.138\linewidth}
      \centering
      \centerline{\includegraphics[width=\linewidth]{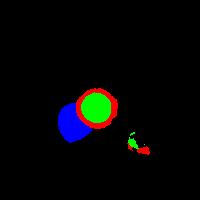}}
    \end{minipage}
    \begin{minipage}[b]{.138\linewidth}
      \centering
      \centerline{\includegraphics[width=\linewidth]{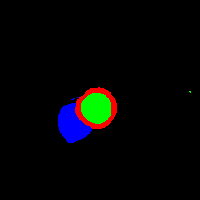}}
    \end{minipage}
    \begin{minipage}[b]{.138\linewidth}
      \centering
      \centerline{\includegraphics[width=\linewidth]{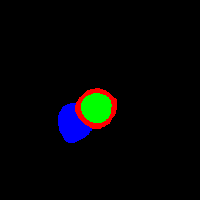}}
    \end{minipage}
    \\
    \begin{minipage}[b]{.138\linewidth}
      \centering
      \centerline{\includegraphics[width=\linewidth]{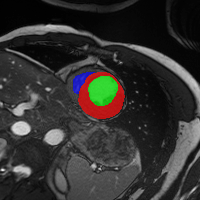}}
      \centerline{(a) image \& GT}\medskip
    \end{minipage}
    \begin{minipage}[b]{.138\linewidth}
      \centering
      \centerline{\includegraphics[width=\linewidth]{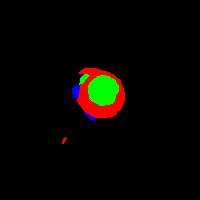}}
      \centerline{(b) MT}\medskip
    \end{minipage}
    \begin{minipage}[b]{.138\linewidth}
      \centering
      \centerline{\includegraphics[width=\linewidth]{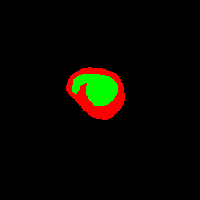}}
      \centerline{(c) UAMT}\medskip
    \end{minipage}
    \begin{minipage}[b]{.138\linewidth}
      \centering
      \centerline{\includegraphics[width=\linewidth]{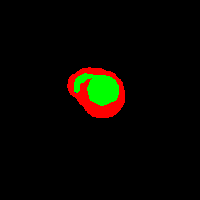}}
      \centerline{(d) ICT}\medskip
    \end{minipage}
    \begin{minipage}[b]{.138\linewidth}
      \centering
      \centerline{\includegraphics[width=\linewidth]{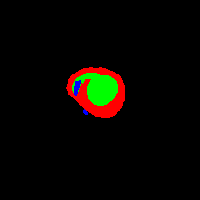}}
      \centerline{(e) CPS}\medskip
    \end{minipage}
    \begin{minipage}[b]{.138\linewidth}
      \centering
      \centerline{\includegraphics[width=\linewidth]{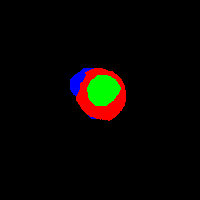}}
      \centerline{(f) URPC}\medskip
    \end{minipage}
    \begin{minipage}[b]{.138\linewidth}
      \centering
      \centerline{\includegraphics[width=\linewidth]{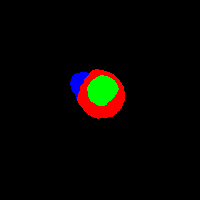}}
      \centerline{(g) EVIL (ours)}\medskip
    \end{minipage}
    \caption{Visual comparison of segmentation results with different methods with 10\%  labeled images.}
    \vspace{-0.4cm}
    \label{fig:my_label}
\end{figure*}

\subsection{Evidential Net (E-Net)}
We follow \cite{sensoy2018evidential} and use cross-entropy loss to make the segmentation probabilities $\boldsymbol p_i$ approach the ground-truth $\boldsymbol y_i$. Notably, the density of $\boldsymbol p_i$ follows the Dirichlet distribution parameterized with $\boldsymbol \alpha_i$. The loss can be formulated as:
\begin{equation}
\begin{aligned}
    \mathcal{L}_{dig} &=\int\left[\sum_{k=1}^{K}-y_{i}^{k} \log \left(p_{i}^{k}\right)\right] D(\boldsymbol{p}_{i} \mid \boldsymbol{\alpha}_{i}) d \boldsymbol{p}_{i} 
    \\
    &=\sum_{k=1}^{K} y_{i}^{k}\left(\psi\left({S}_{i}\right)-\psi\left({\alpha}_{i}^{k}\right)\right),
\end{aligned}
\end{equation}
where $\psi(\cdot)$ is the \textit{digamma} function. By optimizing $\mathcal{L}_{dig}$, the evidence of different classes for positive samples is generated. However, $\mathcal{L}_{dig}$ cannot guarantee that negative samples generate evidence as close as zero. Therefore, Kullback-Leibler (KL) divergence is incorporated into our loss function to penalize the divergence from negative samples, which is defined as:
\begin{equation}
\begin{aligned}
    \mathcal{L}_{K L}
    &=K L\left[
    D\left(\boldsymbol{p}_{i}\mid \tilde{\boldsymbol{\alpha}}_{i}\right)
    \|
    D\left(\boldsymbol{p}_{i} \mid \boldsymbol{1} \right)\right] 
    \\
    &=\log \left(\frac{\Gamma\left(\sum_{k=1}^{K} \widetilde{{\alpha}}_{i}^{k}\right)}{\Gamma(K) \sum_{k=1}^{K} \Gamma\left(\widetilde{\alpha}_{i}^{k}\right)}\right)\\
&+\sum_{k=1}^{K}\left(\widetilde{{\alpha}}_{i}^{k}-1\right)\left[\psi\left(\widetilde{{\alpha}}_{i}^{k}\right)-\psi\left(\sum_{k=1}^{K} \widetilde{{\alpha}}_{i}^{k}\right)\right],
\end{aligned}
\end{equation}
where $\Gamma(\cdot)$ is the \textit{gamma} function, $D(\boldsymbol p_i| \boldsymbol{1})$ is the uniform Dirichlet distribution, and $\tilde{\boldsymbol \alpha}_{i}=\boldsymbol y_i+(1- \boldsymbol y_i) \odot \boldsymbol \alpha_i$.

For segmentation task, the evidence $\boldsymbol e_i$ is obtained with $\boldsymbol x_i$. Then, $\boldsymbol \alpha_i = \boldsymbol  e_i + 1$ is parameterized into the corresponding Dirichlet distribution and the evidential loss is:

\begin{equation}
    \mathcal{L}_{evi} = \mathcal{L}_{dig} + \beta \mathcal{L}_{KL},
\end{equation}
where $\beta$ is a annealing coefficient and is set to $\beta(t)=min(1.0, \frac{t}{0.5 t_{max}})$. $t$ is the current epoch and $t_{max}$ is the total number of training epochs.

As shown in Fig. 2, the Subjective Logic model has two parts, the certain part called belief mass $\boldsymbol b_i$ and the uncertain part $u_i$. The evidential loss generates evidence to reduce the uncertainty. However, since the cross-entropy based evidential loss is based on pixel level, which ignores the relationships between pixels in segmentation task, we use the Dice loss on the certain part and the certain loss is defined as:

\begin{equation}
\begin{aligned}
    \mathcal{L}_{certain}=1 - \frac{2 \sum_{k=1}^K y_i^k \sum_{k=1}^K \hat{p}_i^k}{\sum_{k=1}^K y_i^k + \sum_{k=1}^K \hat{p}_i^k},
\end{aligned}
\end{equation}
where $\hat{\boldsymbol p}_i= softmax(\boldsymbol b_i)$ presents a simplex transformed from the belief mass $\boldsymbol b_i$ with a $softmax$ function.

Then, our overall evidential segmentation loss is defined:
\begin{equation}
    \mathcal{L}_{Eseg} = \mathcal{L}_{evi} + \gamma \mathcal{L}_{certain},
\end{equation}
where $\mathcal{L}_{evi}$ and $\mathcal{L}_{certain}$ denote the evidential loss and the certain loss, respectively. $\gamma$ denotes the weighting parameter, which is set to $1$.
By optimizing $\mathcal{L}_{evi}$, E-Net generates the evidence for positive samples, while reduces the evidence for negative samples. $\mathcal{L}_{certain}$ is leveraged to constrain the relationship between different predicted pixels.

\subsection{EVIL Framework}
The total loss $\mathcal{L}$ for our whole framework contains two components: supervised loss $\mathcal{L}_{sup}$ on labeled data and consistency loss $\mathcal{L}_{con}$ on unlabeled data:
\begin{equation}
    \mathcal{L} = \mathcal{L}_{sup} + \lambda \mathcal{L}_{con},
\end{equation}
where $\lambda$ is the balancing parameter. We use Gaussian ramp-up function $\lambda (t) = \lambda_{max} * e^{-5 \left(1.0 - \frac{t}{t_{max}}\right)^{2}}$ and $\lambda_{max} = 0.1$.

The supervision loss is formulated as:
\begin{equation}
\begin{aligned}
    \mathcal{L}_{sup} = \mathcal{L}_{Eseg} (\mathcal{F}_1(\boldsymbol x), \boldsymbol y) + \mathcal{L}_{Sseg} (\mathcal{F}_2(\boldsymbol x), \boldsymbol y),
\end{aligned}
\end{equation}
where $\mathcal{L}_{Sseg} = \frac{1}{2} (\mathcal{L}_{ce} + \mathcal{L}_{dice} )$ denotes the loss component for S-Net. $\mathcal {L}_{ce}$ and $\mathcal {L}_{dice}$ are the cross-entropy loss and dice loss, respectively.

The pseudo label can be calculated as $Y_1 = argmax(\boldsymbol{b}_i)$ for E-Net and $Y_2 = argmax(\mathcal{F}_2(\boldsymbol x))$ for S-Net. The consistency loss on the unlabeled data is written as:
\begin{equation}
    \mathcal{L}_{con} = \mathcal{L}_{evi}(\mathcal{F}_1(\boldsymbol x), Y_2) + \mathcal{L}_{ce} (\mathcal{F}_2(\boldsymbol x), \mathcal{M} \odot Y_1).
\end{equation}
where $\mathcal{M} = {u} < T$ is the mask to filter out high uncertain results with threshold $T=0.2$. We only use the evidential and cross-entropy losses in consistency loss term due to the mask operation which preserves only the reliable pseudo pixel labels. The consistency loss encourages E-Net to generate potential evidence from S-Net using $\mathcal{L}_{evi}$ and S-Net to learn the reliable pseudo labels using $\mathcal{L}_{ce}$ from E-Net.

\begin{table*}

\centering
\vspace{-1cm}
\caption{The comparison of different methods on ACDC dataset on different semi-supervised labeled data ratio settings.}
\centering
\label{tb3}
\begin{tabular}{@{}c|ccc|ccc|ccc@{}}
\hline
\multicolumn{1}{c|}{\multirow{2}{*}{Method}} & \multicolumn{3}{c|}{10\%}           & \multicolumn{3}{c|}{20\% }           & \multicolumn{3}{c}{30\% } \\
\multicolumn{1}{c|}{}                        & $DSC$ $\uparrow$           & ${HD}_{95}$ $\downarrow$           & $ASD$ $\downarrow$          & $DSC$ $\uparrow$            & ${HD}_{95}$ $\downarrow$            & $ASD$ $\downarrow$           & $DSC$ $\uparrow$         & ${HD}_{95}$ $\downarrow$       & $ASD$ $\downarrow$       \\ \hline
Unet    & 80.05 & 7.41  & 2.38    & 84.90 & 8.94 & 2.52     & 87.07 & 6.61 & 1.95      \\
E-Net (ours)   & 81.05 & 11.17 & 3.26    & 85.68 & 7.39 & 2.12     & 87.45 & 8.12 & 2.23      \\ \hline
MT      & 81.06 & 10.17 & 2.64    & 86.01 & 8.13 & 2.40     & 87.37 & 4.81 & 1.49      \\
UA-MT    & 80.81 & 11.73 & 3.52    & 85.38 & 7.77 & 2.70     & 87.53 & 6.32 & 2.05      \\
ICT     & 83.54 & 8.42  & 2.46    & 85.28 & 5.65 & 1.64     & 87.49 & 8.25 & 2.23      \\
CPS     & 84.70 & 8.25 & 2.35     & 87.47 & 5.98 & 1.74     & 88.21 & 6.49 & 1.90      \\
URPC    & 82.07 & 5.62 & 1.88     & 85.13 & 5.71 & 1.75     & 86.99 & 4.43 & 1.31      \\
EVIL (ours) & \textbf{85.91} & \textbf{3.91} & \textbf{1.36} & \textbf{88.22} & \textbf{4.01} & \textbf{1.21} & \textbf{89.43}  & \textbf{3.84} & \textbf{1.07}           \\ \hline
\end{tabular}
\vspace{-0.2cm}
\end{table*}

\section{Experiment}
\subsection{Experiment Setup}
\label{sec:typestyle}
We evaluate our method on the Automated Cardiac Diagnosis Challenge (ACDC) \cite{bernard2018deep} dataset which contains 200 annotated short-axis cardiac MR-cine images from 100 patients. We leverage 70 patients (140 scans) for training, 10 patients (20 scans) for validation and 20 patients (40 scans) for testing.  All short-axis slices within 3D scans are resized to 256 $\times$ 256 as 2D images. See SSL4MIS \footnote{https://github.com/HiLab-git/SSL4MIS} for more details. For semi-supervised experiments, images from 7 patients, 14 patients and 21 patients are set as labeled ratio 10\%, 20\% and 30\% in the training set, respectively. Standard data augmentation, including random cropping, random rotating, and random flipping, is used to enlarge the training set. Three widely used metrics, Dice Coefficient ($DSC$), Hausdorff Distance 95 (${HD}_{95}$) and  Average Surface Distance ($ASD$) are employed to evaluate the performance of our method.

For the sake of fairness, Unet \cite{ronneberger2015u} is chosen as the backbone in all methods, and
SGD is adopted as the optimizer. The initial learning rate is set to 0.01, and polynomial scheduler strategy is employed to update the learning rate. 
We implement the proposed framework with PyTorch, using a single NVIDIA GTX 1080Ti GPU. The batch size is set to 24, where 12 images are labeled. All methods perform 30000 iterations during training.

\vspace{-0.2cm}
\subsection{Experimental Results}


Several recently proposed semi-supervised segmentation methods are compared, including: Mean-Teacher (MT) \cite{tarvainen2017mean}, Uncertainty-Aware Mean Teacher (UA-MT) \cite{yu2019uncertainty}, Interpolation Consistency Training (ICT) \cite{verma2022interpolation}, Cross Pseudo Supervision (CPS) \cite{chen2021semi}, and Uncertainty Rectified Pyramid Consistency (URPC) \cite{luo2022semi}. For all competing methods, official parameter settings are adopted.

Tab. 1 illustrates the quantitative results on ACDC. The first and second rows list the quantitative results of supervised Unet and E-Net. In different labeled data ratio settings, EVIL outperforms all the other methods. When only 10\% of data are labeled, our method improves $DSC$ by more than 3\% compared with other SOTA uncertainty-aware methods (UAMT and URPC). Moreover, we achieve 4 points improvement in ${HD}_{95}$ and 1 point in $ASD$ compared with CPS. Especially, we can see that the performance of EVIL using 20\% labeled data has surpassed all compared methods using 30\% labeled data.

\begin{table}[]
\vspace{-0.4cm}
\centering
\caption{The comparison of training time.}
\centering
\label{tb3}
\begin{tabular}{@{}c|c|c|c|c@{}}
\hline
Method  & Num & Uncertainty &  Time & Cost      \\ \hline
Unet               & 1 & $\times$ & 0.076 s & -           \\
ICT                & 1 & $\times$ & 0.090 s & + 18.42 \%           \\
URPC               & 1 & $\surd$ & 0.089 s  & + 17.11 \%          \\ 
E-Net (ours)       & 1 & $\surd$  & 0.085 s & + 11.84 \%           \\ \hline
MT                 & 2 & $\times$ & 0.101 s & -           \\
CPS                & 2 & $\times$ & 0.137 s & + 35.64 \%           \\
UA-MT               & 2 & $\surd$ & 0.337 s  & + 233.66\% \\
EVIL (ours)        & 2 & $\surd$ & 0.148 s  & + 46.53 \%        \\\hline
\end{tabular}
\end{table}

Fig. 3 visualizes the segmentation results of two cases using different methods with 10\% labeled data. It is easy to see that the compared methods mis-classify many pixels while EVIL obtains more accurate prediction. As shown in Fig. 4, sampling times affect the uncertainty estimation quality of MC-dropout and our E-Net has best accurate estimation.

Tab. 2 shows the training time with fixed batch size = 24, where `Num', `Uncertainty', `Time', `Cost' denotes the network number, uncertainty-based or not, time consuming, and the additional time consuming cost respectively. We treat Unet as the upper bound of single network method and MT as the baseline of the multi-network framework since it is the fastest method compared to others. Specially, we can see that the proposed method improve significantly without introducing too much computation overhead.
\begin{figure}
    \centering
    \begin{minipage}[b]{.235\linewidth}
      \centering
      \centerline{\includegraphics[width=\linewidth]{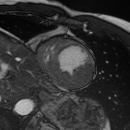}}
      \centerline{(a) Input}\medskip
    \end{minipage}
    \begin{minipage}[b]{.235\linewidth}
      \centering
      \centerline{\includegraphics[width=\linewidth]{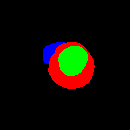}}
      \centerline{(b) Label}\medskip
    \end{minipage}
    \begin{minipage}[b]{.235\linewidth}
      \centering
      \centerline{\includegraphics[width=\linewidth]{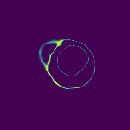}}
        \centerline{(c) Ours}\medskip
    \end{minipage}
    \begin{minipage}[b]{.235\linewidth}
      \centering
      \centerline{\includegraphics[width=\linewidth]{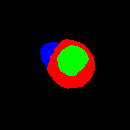}}
        \centerline{(d) EVIL}\medskip
    \end{minipage}
    \\

    \begin{minipage}[b]{.235\linewidth}
      \centering
      \centerline{\includegraphics[width=\linewidth]{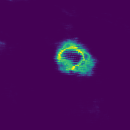}}
      \centerline{(e) S=2}\medskip
    \end{minipage}
    \begin{minipage}[b]{.235\linewidth}
      \centering
      \centerline{\includegraphics[width=\linewidth]{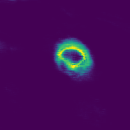}}
      \centerline{(f) S=8}\medskip
    \end{minipage}
    \begin{minipage}[b]{.235\linewidth}
      \centering
      \centerline{\includegraphics[width=\linewidth]{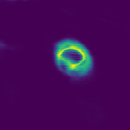}}
      \centerline{(g) S=64}\medskip
    \end{minipage}
        \begin{minipage}[b]{.235\linewidth}
      \centering
      \centerline{\includegraphics[width=\linewidth]{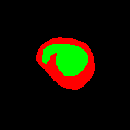}}
      \centerline{(h) UA-MT}\medskip
    \end{minipage}
    \vspace{-0.3cm}
    \caption{Visualization of uncertainty estimation. `S' denotes the MC-dropout sampling times.}
    \vspace{-0.4cm}
\end{figure}


\vspace{-0.2cm}
\section{Conclusion}
\vspace{-0.1cm}
In this paper, we propose a novel uncertainty-aware semi-supervised medical image segmentation framework. The proposed EVIL introduces DST into the consistency regularization training paradigm and achieves fast accurate uncertainty estimation with solid theoretical guarantee.
Extensive experiments demonstrate that EVIL achieves state-of-the-art performance on the widely used ACDC dataset.

\vspace{-0.2cm}
\section{CONFLICTS OF INTEREST}
\vspace{-0.1cm}
The authors declare that they have no conflicts of interest.

\vspace{-0.2cm}
\section{COMPLIANCE WITH ETHICAL STANDARDS}
\vspace{-0.2cm}
This research study was conducted retrospectively using real clinical exams acquired at the University Hospital of Dijon. Ethical approval was not required as confirmed by the license attached with the open access data.

\vspace{-0.2cm}
\section{ACKNOWLEDGEMENT}
\vspace{-0.2cm}
This work was supported in part by the National Natural Science Foundation of China under Grant 62271335; in part by the Sichuan Science and Technology Program under Grant 2021JDJQ0024; and in part by the Sichuan University “From 0 to 1” Innovative Research Program under Grant 2022SCUH0016.
\vspace{-0.2cm}



\bibliographystyle{IEEEbib}
\bibliography{refs}

\begin{thebibliography}{10}

\bibitem{ronneberger2015u}
Olaf Ronneberger, Philipp Fischer, and Thomas Brox,
\newblock ``U-net: Convolutional networks for biomedical image segmentation,''
\newblock in {\em International Conference on Medical Image Computing and
  Computer-Assisted Intervention (MICCAI)}. Springer, 2015, pp. 234--241.

\bibitem{zou2022tbrats}
Ke~Zou, Xuedong Yuan, Xiaojing Shen, Meng Wang, and Huazhu Fu,
\newblock ``Tbrats: Trusted brain tumor segmentation,''
\newblock in {\em International Conference on Medical Image Computing and
  Computer-Assisted Intervention (MICCAI)}. Springer, 2022, pp. 503--513.

\bibitem{zoph2020rethinking}
Barret Zoph, Golnaz Ghiasi, Tsung-Yi Lin, Yin Cui, Hanxiao Liu, Ekin~Dogus
  Cubuk, and Quoc Le,
\newblock ``Rethinking pre-training and self-training,''
\newblock in {\em Advances in Neural Information Processing Systems}, 2020,
  vol.~33, pp. 3833--3845.

\bibitem{feng2020semi}
Zhengyang Feng, Qianyu Zhou, Guangliang Cheng, Xin Tan, Jianping Shi, and
  Lizhuang Ma,
\newblock ``Semi-supervised semantic segmentation via dynamic self-training and
  classbalanced curriculum,''
\newblock {\em arXiv preprint arXiv:2004.08514}, vol. 1, no. 2, pp. 5, 2020.

\bibitem{ibrahim2020semi}
Mostafa~S Ibrahim, Arash Vahdat, Mani Ranjbar, and William~G Macready,
\newblock ``Semi-supervised semantic image segmentation with self-correcting
  networks,''
\newblock in {\em IEEE/CVF Conference on Computer Vision and Pattern
  Recognition (CVPR)}, 2020, pp. 12715--12725.

\bibitem{tarvainen2017mean}
Antti Tarvainen and Harri Valpola,
\newblock ``Mean teachers are better role models: Weight-averaged consistency
  targets improve semi-supervised deep learning results,''
\newblock in {\em Advances in Neural Information Processing Systems}, 2017,
  vol.~30.

\bibitem{chen2021semi}
Xiaokang Chen, Yuhui Yuan, Gang Zeng, and Jingdong Wang,
\newblock ``Semi-supervised semantic segmentation with cross pseudo
  supervision,''
\newblock in {\em IEEE/CVF Conference on Computer Vision and Pattern
  Recognition (CVPR)}. IEEE, 2021, pp. 2613--2622.

\bibitem{ouali2020semi}
Yassine Ouali, C{\'e}line Hudelot, and Myriam Tami,
\newblock ``Semi-supervised semantic segmentation with cross-consistency
  training,''
\newblock in {\em IEEE/CVF Conference on Computer Vision and Pattern
  Recognition (CVPR)}, 2020, pp. 12674--12684.

\bibitem{yu2019uncertainty}
Lequan Yu, Shujun Wang, Xiaomeng Li, Chi-Wing Fu, and Pheng-Ann Heng,
\newblock ``Uncertainty-aware self-ensembling model for semi-supervised 3d left
  atrium segmentation,''
\newblock in {\em International Conference on Medical Image Computing and
  Computer-Assisted Intervention (MICCAI)}. Springer, 2019, pp. 605--613.

\bibitem{wanguncertainty}
Tao Wang, Jianglin Lu, Zhihui Lai, Jiajun Wen, and Heng Kong,
\newblock ``Uncertainty-guided pixel contrastive learning for semi-supervised
  medical image segmentation,''
\newblock in {\em International Joint Conferences on Artificial Intelligence},
  2022.

\bibitem{zheng2021rectifying}
Zhedong Zheng and Yi~Yang,
\newblock ``Rectifying pseudo label learning via uncertainty estimation for
  domain adaptive semantic segmentation,''
\newblock {\em International Journal of Computer Vision}, vol. 129, pp.
  1106--1120, 2021.

\bibitem{audun2018subjective}
Audun Jsang,
\newblock {\em Subjective Logic: A formalism for reasoning under uncertainty},
\newblock Springer, 2018.

\bibitem{Qin2022DECL}
Yang Qin, Dezhong Peng, Xi~Peng, Xu~Wang, and Peng Hu,
\newblock ``Deep evidential learning with noisy correspondence for cross-modal
  retrieval,''
\newblock in {\em ACM International Conference on Multimedia}, 2022, p.
  4948–4956.

\bibitem{sensoy2018evidential}
Murat Sensoy, Lance Kaplan, and Melih Kandemir,
\newblock ``Evidential deep learning to quantify classification uncertainty,''
\newblock in {\em Advances in Neural Information Processing Systems}, 2018,
  vol.~31.

\bibitem{bernard2018deep}
Olivier Bernard, Alain Lalande, Clement Zotti, Frederick Cervenansky, Xin Yang,
  Pheng-Ann Heng, Irem Cetin, Karim Lekadir, Oscar Camara, Miguel
  Angel~Gonzalez Ballester, et~al.,
\newblock ``Deep learning techniques for automatic mri cardiac multi-structures
  segmentation and diagnosis: is the problem solved?,''
\newblock {\em IEEE Transactions on Medical Imaging}, vol. 37, pp. 2514--2525,
  2018.

\bibitem{verma2022interpolation}
Vikas Verma, Kenji Kawaguchi, Alex Lamb, Juho Kannala, Arno Solin, Yoshua
  Bengio, and David Lopez-Paz,
\newblock ``Interpolation consistency training for semi-supervised learning,''
\newblock {\em Neural Networks}, vol. 145, pp. 90--106, 2022.

\bibitem{luo2022semi}
Xiangde Luo, Guotai Wang, Wenjun Liao, Jieneng Chen, Tao Song, Yinan Chen,
  Shichuan Zhang, Dimitris~N Metaxas, and Shaoting Zhang,
\newblock ``Semi-supervised medical image segmentation via uncertainty
  rectified pyramid consistency,''
\newblock {\em Medical Image Analysis}, vol. 80, pp. 102517, 2022.

\end{thebibliography}
\end{document}